\documentclass[letterpaper, 10 pt, conference]{ieeeconf}
\IEEEoverridecommandlockouts

\usepackage{times}

\usepackage[hyphens]{url}  %
\usepackage{graphicx} %
\usepackage{caption} %

\usepackage[style=ieee, maxbibnames=99, citestyle=numeric-comp]{biblatex}
\usepackage{multicol}
\usepackage{booktabs}

\graphicspath{{../}{./}{./figures/}}
\usepackage[utf8]{inputenc}

\usepackage{svg}
\usepackage{amsmath}
\usepackage[font=scriptsize,labelfont=bf]{caption}
\usepackage{graphicx}
\usepackage{xcolor}
\usepackage[justification=centering]{subfig}
\usepackage{amsmath}
\usepackage{amsfonts}
\usepackage{amssymb}
\usepackage{mathtools}
\usepackage{todonotes}
\let\labelindent\relax
\usepackage{enumitem}
\usepackage{comment}
\usepackage[per-mode=symbol]{siunitx}
\DeclareSIUnit\minute{min}
\usepackage{tikzpagenodes}
\usepackage{algorithm}
\usepackage[noend]{algpseudocode}
\newcommand\NoDo{\renewcommand\algorithmicdo{}}

\newcommand\NoThen{\renewcommand\algorithmicthen{}}

\floatname{algorithm}{Algorithm}

\usepackage{geometry}
\makeatletter
\let\NAT@parse\undefined
\makeatother
\usepackage[hidelinks]{hyperref}

\renewcommand\footnotemark{}
\renewcommand\footnoterule{}
\makeatletter
\def\footnoterule{\kern-3\p@
  \hrule \@width 2in \kern 2.6\p@} 
\makeatother

\begin{document}

\bstctlcite{IEEEexample:BSTcontrol}


\title{Sequential Bayesian Optimization for \\ Adaptive Informative Path Planning with Multimodal Sensing 

}

\author{
    Joshua Ott\textsuperscript{\rm 1}\thanks{\textsuperscript{\rm 1}Department of Aeronautics \& Astronautics, Stanford University \newline \indent (\{joshuaott, mykel\}\!@stanford.edu).},
    Edward Balaban\textsuperscript{\rm 2}\thanks{\textsuperscript{\rm 2}NASA Ames Research Center (edward.balaban@nasa.gov).},
    and Mykel J. Kochenderfer\textsuperscript{\rm 1}
}

\maketitle

\renewcommand{\thefootnote}{\fnsymbol{footnote}}

\begin{abstract}
Adaptive Informative Path Planning with Multimodal Sensing (AIPPMS) considers the problem of an agent equipped with multiple sensors, each with different sensing accuracy and energy costs. The agent's goal is to explore the environment and gather information subject to its resource constraints in unknown, partially observable environments. Previous work has focused on the less general Adaptive Informative Path Planning (AIPP) problem, which considers only the effect of the agent's movement on received observations. The AIPPMS problem adds additional complexity by requiring that the agent reasons jointly about the effects of sensing and movement while balancing resource constraints with information objectives. We formulate the AIPPMS problem as a belief Markov decision process with Gaussian process beliefs and solve it using a sequential Bayesian optimization approach with online planning. Our approach consistently outperforms previous AIPPMS solutions by more than doubling the average reward received in almost every experiment while also reducing the root-mean-square error in the environment belief by 50\%. We completely open-source our implementation to aid in further development and comparison.\footnote[2]{\url{https://github.com/sisl/SBO_AIPPMS}}

\end{abstract}

\IEEEpeerreviewmaketitle

\section{Introduction}\label{sec:intro}
In many real-world applications such as planetary exploration, search-and-rescue, and environmental monitoring, an autonomous agent must explore an unknown environment with partial observability. The agent is equipped with multiple sensors that allow it to take noisy observations of the environment. In addition, the agent has limited time and energy resources requiring that it balance exploring the environment and exploiting its knowledge of the environment based on its current belief. This problem is referred to as Adaptive Informative Path Planning with Multimodal Sensing (AIPPMS) \cite{choudhury2020adaptive}. The objective of the AIPPMS problem is to produce an adaptive decision making strategy to guide the agent to the goal location while balancing exploration with exploitation and respecting the resource constraints. 

\begin{figure}[t]
\centering
    {\includegraphics[width=1\columnwidth]{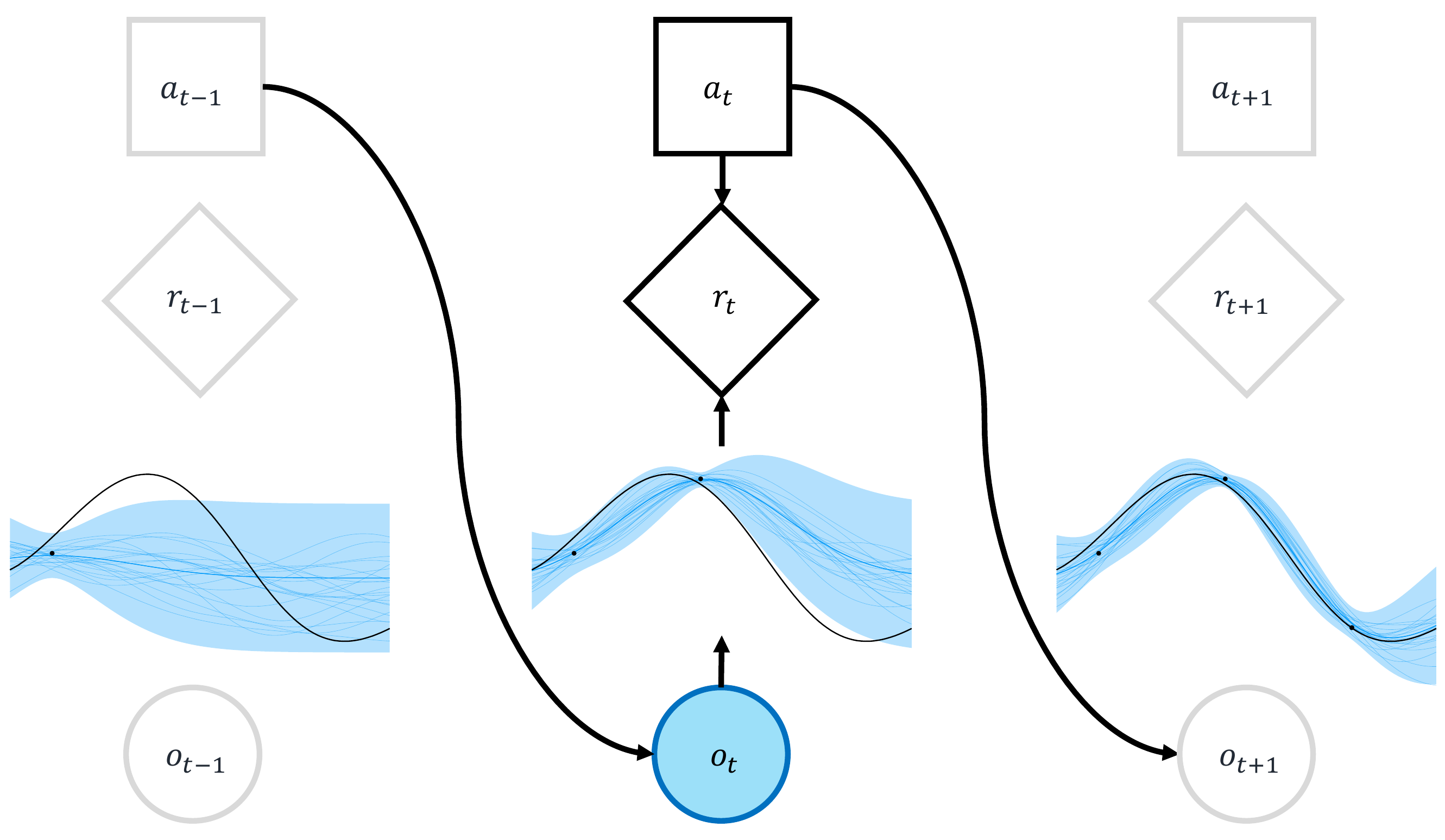}}    
  \caption{The belief MDP framework where the belief-state is a Gaussian process. Upon taking an action, the agent receives a noisy observation of the environment, updates its belief-state, and receives a reward based on the belief-state and action taken. Sequential Bayesian optimization aims to find the best place to take samples of an unknown function accounting for the sequential constraints associated with the agent's movement and energy costs. For clarity, we illustrate a one-dimensional Gaussian process, but our method is not restricted to one dimension. } \label{fig:front_page_fig}
  \vspace{-7mm}
\end{figure}

Many real-world scenarios involve agents equipped with multiple sensors that cannot be simultaneously active due to energy, sensor interference, or movement constraints. Consider an agent with multimodal sensing capabilities. In this context, multimodal refers to the existence of multiple sensors with potentially different levels of accuracy and cost associated with each sensor; for example, a rover equipped with a suite of sensors (e.g., CCD cameras, radar, mass spectrometers, and drills). Some of these sensors, such as the cameras, require small amounts of energy but produce less reliable measurements because they are unable to fully observe the subsurface environment. Conversely, other sensors, such as the drill, require very large amounts of energy but produce almost perfect measurements of the subsurface environment. The same trade-off between sensor efficiency and sensor accuracy can be found in many other applications such as ocean monitoring, mineral prospecting, aerial surveillance, and wildfire mapping.

The informative path planning (IPP) problem is NP hard \cite{meliou2007nonmyopic} and involves an agent planning a path without accounting for noisy observations of the environment. The adaptive problem extends the IPP problem by seeking a policy that changes in response to received observations \cite{singh2009nonmyopic, hollinger2013active, hitz2014fully, lim2016adaptive, girdhar2016modeling}. The AIPPMS problem further extends the adaptive problem by reasoning over multiple sensing modalities with different sensing costs corresponding to sensor accuracy and movement costs \cite{choudhury2020adaptive}.

Both the adaptive informative path planning (AIPP) and non-adaptive IPP problems have been widely studied and a variety of approaches have been proposed \cite{toussaint2014bayesian, popovic2020informative, duecker2021embedded, ott2022riskaware, morere2018continuous, ruckin2022adaptive}. Most of the previous work has focused on the case where the agent is equipped with a single sensor. In this work, we are interested in the case where the agent can choose between multiple sensing modalities. This extension adds additional complexity by requiring that the agent reasons about the cost-benefit trade-off associated with the different sensing modalities. Choudry et al. considered multiple sensing modalities \cite{choudhury2020adaptive}. We build upon this work by formulating the problem as a belief Markov decision process (MDP) as shown in Fig. \ref{fig:front_page_fig}. We solve the AIPPMS problem through a sequential Bayesian optimization approach using Monte Carlo tree search with Double Progressive Widening (MCTS-DPW) and belief-dependent rewards \cite{kaelbling1998planning, couetoux2011continuous, kochenderfer2022algorithms}. We directly compare our results with those of Choudry et al. and show a significant improvement in performance by taking sequential actions to maximize the approximate mutual information based on the current Gaussian process belief.

The key contributions of this work are:
\begin{enumerate}
    \item The formulation of the AIPPMS problem as a belief MDP where the world belief-state is represented as a Gaussian process. 

    \item The extensive evaluation of our method in simulation, along with directly comparing to previous AIPPMS methods. We compare our results on the Information Search RockSample problem introduced by He et al. \cite{he2011efficient} and adapted by Choudry et al. \cite{choudhury2020adaptive}. We also introduce a new benchmark for AIPPMS problems: the Rover Exploration problem.

    \item The release of our implementation as an open-source software package for use and further development by the community.
\end{enumerate}

The remainder of this paper is organized as follows. Section \ref{sec:relatedwork} discusses related work. Section \ref{sec:background} provides preliminary background on Gaussian processes and partially observable Markov decision processes (POMDP). Section \ref{sec:methods} formulates the AIPPMS problem as a belief MDP and describes our approach for solving the problem. Section \ref{sec:results} presents our results on two AIPPMS problems and compares our method to Choudry et al. and other baseline methods \cite{choudhury2020adaptive}. Section VI concludes our work.

\section{Related Work}\label{sec:relatedwork}

A common approach for solving the AIPP problem is to represent the world belief as a Gaussian process. Gaussian processes are distributions over functions and are able to represent spatial correlation among observations while quantifying uncertainty in the underlying function \cite{kochenderfer2019algorithms}. Marchant et al. formulate the AIPP problem as a POMDP and solve it using sequential Bayesian optimization through MCTS Upper Confidence bound for Trees (MCTS-UCT) \cite{marchant2014sequential}. Extensions of this work include modifying the reward function to achieve a monitoring behavior that exploits high-gradient areas \cite{morere2017sequential} and further reasoning over continuous action spaces through Bayesian optimization \cite{morere2018continuous}. Rayas Fernandez et al. use Partially Observable Monte-Carlo Planning (POMCP) with Gaussian process beliefs to estimate quantiles of the underlying world state \cite{fernandez2022informative}. Their approach suggests sample locations for a team of scientists to visit once it has completed its exploration of the environment.  

More recent methods have used Gaussian Markov random fields (GMRF) to model the world belief. GMRFs exploit the spatial Markov property to enforce conditional independence of non-neighboring lattice nodes in the world belief \cite{duecker2021embedded}. This approach avoids the $\mathcal{O}(n^3)$ inference requirement that is common among Gaussian processes, but introduces additional assumptions through the conditional independence of non-neighboring nodes. Rückin et al. combines tree search with an offline-learned neural network predicting informative sensing actions \cite{ruckin2022adaptive}.

In these works, the only decision is the next location to visit, and energy is only depleted by movement. Choudry et al. introduced the AIPPMS problem where the agent must make decisions about where to visit and what sensors to use \cite{choudhury2020adaptive}. Greater energy cost is incurred for using more accurate sensors. Choudry et al. formulate the AIPPMS problem as a POMDP and use POMCP to produce a solution. 

We extend this work by formulating the AIPPMS problem as a belief MDP and use belief-dependent rewards to penalize uncertainty in the belief through sequential Bayesian optimization-based rollouts with progressive widening. 



\vspace{-3mm}
\section{Background}\label{sec:background}
\subsection{Gaussian Processes}
A Gaussian process is a distribution over functions that can be used to predict an underlying function $f$ given some previously observed noisy measurements at location $\textbf{x}$. That is, $y = f(\textbf{x}) + z$ where $f$ is deterministic but $z$ is zero-mean Gaussian noise, $z \sim \mathcal{N}(0, \nu)$. The distribution of functions conditioned on the previously observed measurements  $\boldsymbol{\hat{y}} \mid \boldsymbol{y}, \nu \sim \mathcal{N}(\boldsymbol{\mu^{\ast}}, \boldsymbol{\Sigma^{\ast}})$ is given by: \footnotesize
\begin{gather} \label{eq:gp_update}
    \boldsymbol{\mu^{\ast}} = \boldsymbol{m}(X^{\ast})
    \newline + \boldsymbol{K}(X^{\ast}, X)(\boldsymbol{K}(X, X) + \nu \boldsymbol{I})^{-1} (\boldsymbol{y} - \boldsymbol{m}(X)) \\
    \nonumber \boldsymbol{\Sigma^{\ast}} = \boldsymbol{K}(X^{\ast}, X^{\ast}) - \boldsymbol{K}(X^{\ast}, X)(\boldsymbol{K}(X,X) + \nu \boldsymbol{I})^{-1} \boldsymbol{K}(X, X^{\ast})
\end{gather} \normalsize where $X$ is the set of previously measured locations, $X^{\ast}$ is the set of locations we wish to predict the values $\boldsymbol{\hat{y}}$, $\boldsymbol{m}(X)$ is the mean function and $\boldsymbol{K}(X,X')$ is the covariance matrix constructed with the kernel $k(\boldsymbol{x},\boldsymbol{x'})$ \cite{kochenderfer2019algorithms}.


    
\vspace{-2mm}
\subsection{POMDPs}
A POMDP is a problem formulation for sequential decision making under uncertainty and is represented as a tuple $(\mathcal{S}, \mathcal{A}, O, T, \mathcal{Z}, R, \gamma)$, where $\mathcal{S}$, $\mathcal{A}$, and $\mathcal{O}$ denote the state, action, and observation spaces, respectively. At each time step, the agent takes an action $a \in \mathcal{A}$ to move from state $s \in \mathcal{S}$ to $s' \in \mathcal{S}$ and then receives an observation $o \in O$. The transition function $T(s,a,s') = p(s' \mid s, a)$ specifies the probability of transitioning to $s'$ given that action $a$ was taken from state $s$. Similarly, the observation function $Z(s', a, o) = p(o \mid s', a)$ specifies the probability of observing $o$ given that the agent took action $a$ and transitioned to state $s'$. The belief over system states, $b(s)$, is a probability distribution over $\mathcal{S}$. The agent receives a reward specified by $R(s,a)$ for taking action $a$ in state $s$.
\vspace{-2mm}
\subsection{Belief MDP}
Any POMDP can be viewed as an MDP that uses beliefs as states, also known as a belief MDP \cite{kaelbling1998planning}. The state space of a belief MDP is the set of all beliefs and is therefore continuous. The action space is identical to that of the POMDP. The reward function for a belief MDP is dependent on the belief and action taken. Formulating the problem as a belief MDP allows one to directly access the belief-state in the reward function which is useful in information-gathering tasks. 


\section{Methods}\label{sec:methods}
\vspace{-4mm}
\subsection{Belief MDP Formulation} \label{sec:bmdp_formulation}
\vspace{-3mm}
We formulate and solve the AIPPMS problem as a belief MDP in order to jointly reason about the effect of movement and multimodal sensing actions. The world environment is represented by a location graph $\mathcal{G} = (\mathcal{V}, \mathcal{E})$, where the nodes $\mathcal{V}$ correspond to the set of all locations the agent can visit. The agent starts at $v_s$ and needs to reach a specified goal $v_g$ while remaining within its total cost budget $B$. The agent is equipped with a set of sensors with each sensor having a different cost and level of accuracy. We use notation consistent with previous work on the AIPP problem \cite{singh2009nonmyopic, choudhury2020adaptive}.

\textbf{Belief State Space $\mathcal{B}$}: The belief-state at time $t$ is defined as $b_t = (v_t, \Delta e_t, GP(m(\boldsymbol{x}), k(\boldsymbol{x}, \boldsymbol{x'})))$ where $v_t$ is the agent's current location and $\Delta e_t = B - \sum_{i=1}^{t-1} C(b_i, a_i)$ is the remainder of the cost budget. The cost associated with taking action $a_i$ from belief-state $b_i$ is $C(b_i, a_i)$. Different costs are incurred for movement and sensing actions. Similarly, selecting a sensing action with greater accuracy will incur more cost than a sensing action with less accuracy. Note that the cost budget is different from the reward and is meant to track the time and energy resources of the agent. The problem terminates when the agent has too few resources left to take any actions, which occurs when $\Delta e < \text{min}_a C(b_t,a)$. If the agent is not at the goal location $v_g$ when this happens it receives $-\infty$ reward. The belief over the world state $\mathcal{X}_\mathcal{V}$ is represented as a Gaussian process where $\mathcal{X}_\mathcal{V} \sim GP(m(\boldsymbol{x}), k(\boldsymbol{x}, \boldsymbol{x'}))$. 
    
\textbf{Action Space $\mathcal{A}$:} The action space is 
$\text{neigh}(v_t) \cup S$ where $\text{neigh}(v_t)$ is all of the neighboring locations to $v_t$ and $S$ is the space of sensing actions.
    
\textbf{Transition Model $T$:} The transition function is fully deterministic for $v_t$ and $\Delta e_t$. If $a_t$ is a movement action to a neighboring node, then $v_{t+1} = a_t$ and $\Delta e_{t+1} = \Delta e_{t} - C(b_t, a_{t+1})$. If $a_t$ is a sensing action, then $v_{t+1} = v_t$ and the remaining cost budget will change accordingly. The transition for the world belief involves an update of the current belief-state based on the result of any sampling actions taken. If a sampling action is taken, a measurement $y_t$ will be received at location $v_t$ and the Gaussian process update equations provided in Eq. \eqref{eq:gp_update} can be used to obtain the posterior distribution $\boldsymbol{\hat{y}} \mid \boldsymbol{y_t}$ where $v_t$ is appended to the set of previously measured locations $M_t$, and $y_t$ is appended to the set of previously received measurements $\boldsymbol{y_t}$.
    
\textbf{Reward Function $R$:} 
The reward function is made up of two key elements. The first is the expected reward of interacting with the environment based on our current belief: $\sum_s R(s,a)b(s)$. The second is based on the goal of minimizing the uncertainty in the world belief. To do so, we use the mutual information. We have that $\mathcal{X}_{\mathcal{V}}$ refers to the set of random variables at all locations in the environment and $M_t$ contains all of the locations that have been observed by the agent at time $t$. The mutual information between the locations $M_t$ and the rest of the world $\mathcal{X}_{\mathcal{V}}$ given previous measurement locations $M_{t-1}$ is: 

\small
\begin{align} \label{eq:mutual_information}
    I(\mathcal{X}_{M_t} ; \mathcal{X}_{\mathcal{V} \mid M_{t-1}} ) = \mathcal{H}( \mathcal{X}_{\mathcal{V} \mid M_{t-1}} ) - \mathcal{H}( \mathcal{X}_{\mathcal{V} \mid M_{t-1}} \mid \mathcal{X}_{M_t}) \\
    \nonumber = \mathcal{H}( \mathcal{X}_{\mathcal{V}} \mid \mathcal{X}_{M_{t-1}} ) - \mathcal{H}( \mathcal{X}_{\mathcal{V}} \mid \mathcal{X}_{M_{t}} )
\end{align} \normalsize where $\mathcal{H}( \mathcal{X}_{\mathcal{V}} \mid \mathcal{X}_{M_{t}})$ is the conditional entropy. For a Gaussian process, the conditional entropy is 
\begin{gather}
    \mathcal{H}(\mathcal{X}_{\mathcal{V}} \mid \mathcal{X}_{M_t}) = \frac{1}{2} \log \lvert \boldsymbol{\Sigma^{\ast}_{M_t}} \rvert + \frac{D}{2} (1 + \log(2\pi)).
    \label{eq:analytical_cond_entropy}
\end{gather} Therefore, Eq. \eqref{eq:mutual_information} can be rewritten as 
\begin{gather} \label{eq:mutual_information_rewritten}
    I(\mathcal{X}_{M_t} ; \mathcal{X}_{\mathcal{V} \mid M_{t-1}} ) = \frac{1}{2} \log \lvert \boldsymbol{\Sigma^{\ast}_{M_{t-1}}} \rvert - \frac{1}{2} \log \lvert \boldsymbol{\Sigma^{\ast}_{M_t}} \rvert
\end{gather} where we drop the constant term from Eq. \eqref{eq:analytical_cond_entropy}. Mutual information was proposed as a sampling criterion by Caselton and Zidek \cite{caselton1984optimal}. It seeks to find sensor placements that are most informative about unsensed locations. This optimization criterion directly measures the effect of sensor placements on the posterior uncertainty of the Gaussian process. To avoid computing the determinant at every step during the rollout, we approximate Eq. \eqref{eq:mutual_information_rewritten} with 
\begin{gather} \label{eq:mutual_information_approx}
    I(\mathcal{X}_{M_t} ; \mathcal{X}_{\mathcal{V} \mid M_{t-1}} ) \approx \text{Tr} \left( \boldsymbol{\Sigma^{\ast}_{M_{t-1}}} \right) - \text{Tr} \left( \boldsymbol{ \Sigma^{\ast}_{M_{t}}} \right),
\end{gather} where the trace of the covariance matrix corresponds to the total variance in the world belief-state. As shown by Krause et al., minimizing the conditional entropy in Eq. \eqref{eq:analytical_cond_entropy} is equivalent to sampling at locations of highest entropy. For a Gaussian process, this is equivalent to sampling at locations with the greatest predictive variance \cite{krause2008near}. Similarly, sampling at locations with the greatest predictive variance is equivalent to sampling at locations that decrease the predictive variance the most. The predictive variance at each location in $\mathcal{X}_{\mathcal{V}}$ is given by the diagonals of the posterior covariance matrix $\boldsymbol{\Sigma^{\ast}}$. Therefore, we seek to take the sequence of actions that maximally reduce the predictive variance of the world belief-state, and thereby approximately maximize the mutual information between the new sample locations and the rest of the world belief-state given the previous measurement locations. The reward function is then given by:
\begin{equation} \label{eq:reward}
        R(b, a, b') = \sum_s R(s,a)b(s) + \lambda I(\mathcal{X}_{M_t} ; \mathcal{X}_{\mathcal{V} \mid M_{t-1}} )
\end{equation} where $\lambda$ weights the mutual information uncertainty reduction against the expected reward.

\subsection{Online Sequential Bayesian Optimization Solution}
As opposed to classical Bayesian optimization methods that greedily reason about the next sample location, we use a tree search approach to reason about the agent's actions several steps into the future. To solve the belief MDP, we use MCTS-DPW, which accommodates continuous state and action spaces. MCTS-DPW with belief-states is very similar to the PFT-DPW algorithm described by Sunberg and Kochenderfer \cite{sunberg2018online}. The main difference being they use a particle representation of the belief and apply particle filter belief updates in the simulate function of the algorithm. Marchant et al. outlined a sequential Bayesian optimization approach using MCTS but did not make use of progressive widening for continuous state and action spaces \cite{marchant2014sequential}. 

MCTS is a widely studied algorithm for online decision making \cite{browne2012survey}. MCTS constructs a tree of state and action nodes using a generative model $G$ and estimates the state-action value function $Q(ba)$ at each of the action nodes. Nodes are expanded using the upper confidence bound $UCB(b,a) = Q(ba) + c\sqrt{\frac{\log N(b)}{N(ba)}}$ where $N(ba)$ is the number of times the action node has been visited, $N(b) = \sum_{a \in \mathcal{A}} N(ba)$, and $c$ is a parameter that determines the weight given to exploration in the tree. 

\begin{algorithm}[t]
\scriptsize
\caption{\scriptsize Monte Carlo Tree Search with Double Progressive Widening}\label{alg:cap}
\begin{algorithmic}
\Function{Plan}{$b$}
  \NoDo
  \For{$i \in 1:n$}
        \State \textsc{Simulate}($b, d_{\text{max}})$
  \EndFor
  \State \Return $\underset{a}{\text{argmax}} Q(ba)$
\EndFunction

\Function{ActionProgWiden}{$b$}
    \NoThen
    \If{$|C(b)| \leq k_a N(b)^{\alpha_a}$}
        \State $a \leftarrow \textsc{NextAction}(b)$ \Comment{feasible actions only}
        \State $C(b) \leftarrow C(b) \cup \{a\}$
    \EndIf
    \State \Return $\underset{a \in C(b)}{\text{argmax }} Q(ba) + c\sqrt{\frac{\log N(b)}{N(ba)}}$
\EndFunction

\Function{Simulate}{$b,d$}
    \If{$d = 0$}
        \State \Return $0$
    \EndIf
    \State $a \leftarrow \textsc{ActionProgWiden}(b)$
    \If {$|C(ba)| \leq k_s N(ba)^{\alpha_s}$}
        \State $(b',r) \sim G_{GP}(b,a)$
        \State $C(ba) \leftarrow C(ba) \cup \{(b',r) \}$
        \State $q \leftarrow r + \gamma \textsc{Rollout}(b', d-1)$
    \Else
        \State $(b',r) \leftarrow \text{sample uniformly from } C(ba)$
        \State $q \leftarrow r + \gamma \textsc{Simulate}(b', d-1)$
    \EndIf
    \State $N(b) \leftarrow N(b)+1$
    \State $N(ba) \leftarrow N(ba)+1$
    \State $Q(ba) \leftarrow Q(ba)+\frac{q-Q(ba)}{N(ba)}$
    \State \Return $q$
\EndFunction
\end{algorithmic}
\end{algorithm}

\normalsize
When the state and action spaces are large or continuous, MCTS will produce trees that are very shallow. 
With MCTS-DPW, the number of new children sampled from any node in the search tree is limited by DPW using the parameters $k_s$, $\alpha_s$, $k_a$, and $\alpha_a$. The number of children of a node in the tree is limited to $kN^{\alpha}$ where $N$ is the number of times the node has been visited and $k$ and $\alpha$ are hyper-parameters \cite{couetoux2011adding}. When the number of state nodes is greater than $kN^{\alpha}$, one of the previously generated states is chosen instead of simulating a new state transition. The belief-state implementation of MCTS-DPW is provided in Algorithm 1 following the notation of Sunberg and Kochenderfer, where $G_{GP}$ denotes a Gaussian process belief update \cite{sunberg2018online}.

\floatsep 1\baselineskip plus  0.2\baselineskip minus  0.2\baselineskip
\textfloatsep 1.7\baselineskip plus  0.2\baselineskip minus  0.4\baselineskip

Similar to Choudhury et al., we apply action pruning so that we only consider feasible actions during the tree search. A feasible action is one that transitions the agent to a feasible state where a feasible state is a state in which the agent is still able to reach the goal within the specified resource constraints. That is, belief-state $b_t$ is feasible if $\Delta e_t > C_G(v_t,v_g)$ where $C_G(v_t,v_g)$ is the cost along the shortest path from the agents current location to the goal location. We use the POMDPs.jl framework to implement the preceding algorithms in our experiments \cite{egorov2017pomdps}.

\section{Results}\label{sec:results}
\begin{figure}[t]
\centering
    {\includegraphics[width=0.5\textwidth]{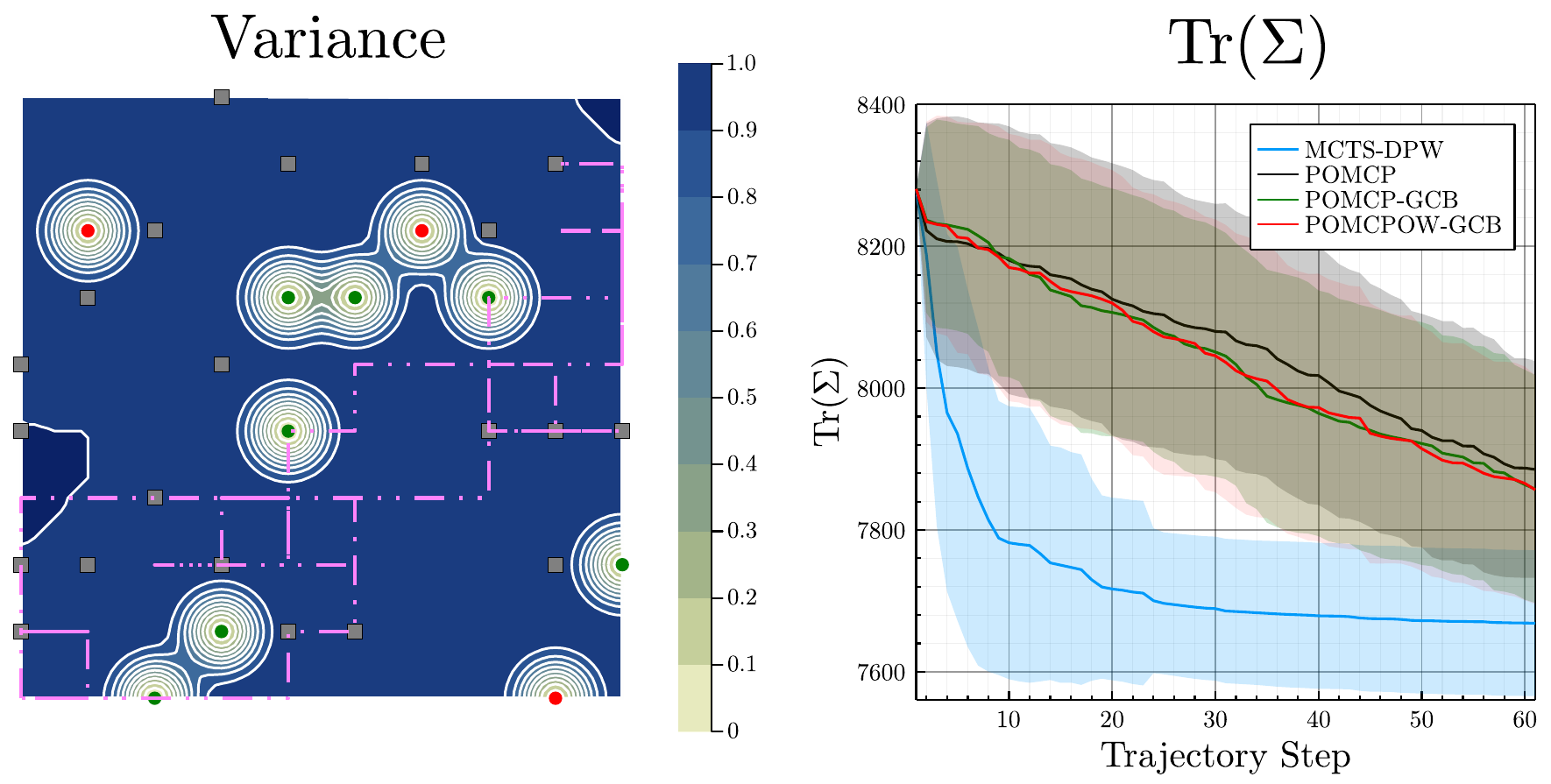}}    
  \caption{Results from the ISRS problem. The left image shows the posterior variance of the Gaussian process belief using MCTS-DPW. Bad rocks are shown in red, good rocks in green, and beacons in gray. The agent's trajectory is shown in pink. The right plot shows $\text{Tr}(\Sigma)$ along the agent's trajectory. Note that for even comparison we evaluated $\text{Tr}(\Sigma)$ using the same Gaussian process setup, even though MCTS-DPW was the only method to use Gaussian process beliefs online.}\label{fig:isrs_results}
  \vspace{-7mm}
\end{figure}

\subsection{Information Search RockSample}
We first evaluate our method on the Information Search RockSample (ISRS) problem introduced by He et al. \cite{he2011efficient} and adapted by Choudry et al. \cite{choudhury2020adaptive}. ISRS is a variation of the classic RockSample problem. The agent must move through an environment represented as an $n \times n$ grid. Scattered throughout the environment are $k$ rocks with at most one rock in each grid cell. Only some of the rocks are considered to be `good,' meaning that they have some scientific value. The agent receives $+10$ reward for visiting a good rock and $-10$ reward for visiting a bad rock. Once a good rock is visited, it becomes bad. The positions of the agent and rocks are known apriori, but visiting a rock is the only way to reveals its true state. 

\begin{table*}[t]
\resizebox{\textwidth}{!}{
\Large 
\begin{tabular}{ cccccccccccccccc }
\toprule
     & &\multicolumn{4}{c}{$p=0.5$} & &\multicolumn{4}{c}{$p=0.75$} & & \multicolumn{4}{c}{$p=1.0$} \\
    \cline{3-6}
    \cline{8-11}
    \cline{13-16} \\
    $k$ & $b$ & POMCP & POMCP-GCB &  POMCPOW-GCB & MCTS-DPW & & POMCP & POMCP-GCB & POMCPOW-GCB & MCTS-DPW & & POMCP & POMCP-GCB & POMCPOW-GCB & MCTS-DPW  \\

    \hline 
    $10$ & $10$ & $-24.0$ & $-18.0$ & $-20.2$  & $\boldsymbol{5.8}$ & & $-1.6$ & $4.0$ & $11.0$ & $\boldsymbol{13.0}$ & & $31.4$ & $35.8$ & $\boldsymbol{42.8}$ & $26.2$ \\
    $10$ & $25$ & $-26.0$ & $-16.4$ & $-20.4$ & $\boldsymbol{12.2}$ & & $-1.4$ & $5.8$ & $7.4$ & $\boldsymbol{21.6}$ & & $32.8$ & $\boldsymbol{39.2}$ & $38.0$ & $33.6$ \\
    $25$ & $10$ & $-60.0$ & $-54.2$ & $-61.0$ & $\boldsymbol{15.2}$ & & $9.8$ & $17.8$ & $13.6$ & $\boldsymbol{30.4}$ &  & $80.6$ & $\boldsymbol{95.0}$ & $77.2$ & $61.4$ \\
    $25$ & $25$ & $-61.4$ & $-40.4$ & $-47.8$ & $\boldsymbol{18.0}$ & & $8.4$ & $15.0$ & $9.2$ & $\boldsymbol{41.2}$  & & $93.4$ & $\boldsymbol{100.8}$ & $89.0$ & $73.4$ \\
\bottomrule
\end{tabular}}
\caption{Comparison of the three different methods as the distribution of good rocks, number of rocks, and number of beacons are varied. Each reported value is the average of 50 different simulation runs. Note that we report different values from Choudhury et al. since we have included a $-10$ reward for visiting a bad rock whereas they do not. }
\vspace{-7mm}
\label{table:isrs_results}
\end{table*}

There are also $b$ beacons scattered throughout the environment. Upon reaching a beacon location, the agent has the option to take a sensing action where it will receive observations about the state of the nearby rocks. The fidelity of the observation decreases with increasing distance from the beacon location. There are multiple sensing modalities available to the agent with a higher cost for choosing the more accurate sensing modality. Moving between adjacent cells also expends energy cost. The agent's goal is to visit as many good rocks and as few bad rocks as possible while returning to the origin without exceeding its resource budget. 

Similar to the approach taken by Choudhury et al., we focus on the relationship between the number of rocks $k$, the number of beacons $b$, and the relative proportion of good rocks, based on the independent Bernoulli probability $p$ of a rock being good. We vary these three parameters in our experiments while keeping the others fixed. We directly compare our approach with the POMCP and POMCP-GCB approach used by Choudhury et al. \cite{choudhury2020adaptive}. In addition, we compare against the POMCPOW algorithm introduced by Sunberg and Kochenderfer \cite{sunberg2018online}, where we use the Generalized Cost Benefit (GCB) algorithm in the rollouts. In the remainder of the paper, we refer to this algorithm as POMCPOW-GCB. Choudhury et al. formulates the problem as a POMDP with discrete beliefs over whether a rock is good or bad and updates the belief as observations are received. They solve the problem using POMCP with random rollouts and a variation of POMCP with a GCB rollout policy. The main difference is that we formulate the problem as a belief MDP, represent the belief over the world state as a Gaussian process, and use the mutual information $I(\mathcal{X}_{M_t} ; \mathcal{X}_{\mathcal{V} \mid M_{t-1}} )$ in our MCTS-DPW solver. We refer the reader to Choudhury et al. for further details \cite{choudhury2020adaptive}.

Fig. \ref{fig:isrs_results} shows the variance of the world belief after the agent has finished exploring the environment. The figure also shows $\text{Tr}(\Sigma)$ as the agent moves along its trajectory. Table \ref{table:isrs_results} compares the results as the $k$, $b$, and $p$ parameters are varied across 50 different simulation runs. We see that the belief MDP formulation with Gaussian process beliefs and MCTS-DPW is able to significantly outperform POMCP, POMCP-GCB, and POMCPOW-GCB in situations where there are both good and bad rocks present. Intuitively, this makes sense because MCTS-DPW is able to take information-gathering actions to reduce the uncertainty in the world belief-state. This encourages the agent to take sensing actions to become more certain about the true value of the rocks before sampling them. POMCP does not explicitly reason about information-gathering actions. Instead, as the random rollouts are performed and more sensing actions are taken, the agent will become more certain about the state of the environment, thereby indirectly benefiting from sensing. With the GCB rollout policy, sensing actions are incentivized to maximize in expectation the mode of the belief-state \cite{choudhury2020adaptive}. This incentive leads the agent to take actions that have a greater impact on the difference of the belief mode; however, it is not as robust as reasoning about the mutual information through sequential Bayesian optimization as shown in Table 1. Other work by Fischer and Tas has observed similar performance when information-gathering actions are a key to successful execution \cite{fischer2020information}. 

When only good rocks are present $(p=1.0)$, MCTS-DPW does not outperform the other methods. This is because MCTS-DPW still initially prioritizes sensing actions to reduce its uncertainty in the environment belief, whereas the other methods greedily head toward the rocks with fewer sensing actions. Intuitively, this makes sense because sensing actions are not as valuable in homogeneous environments where POMCP, POMCP-GCB, and POMCPOW-GCB learn that they will receive reward for visiting any rock location. 

\begin{figure*}[t]
\centering
    {\includegraphics[width=1\textwidth]{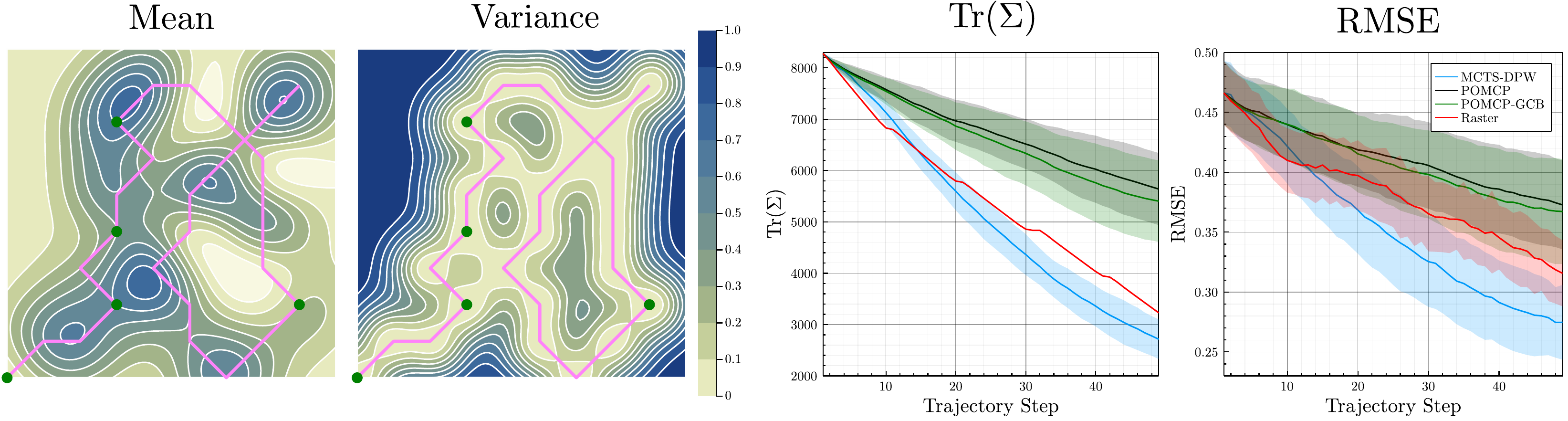}}    
  \caption{From left to right: contour plot showing the posterior mean and variance of the Gaussian process belief after the agent has reached the goal state. The agent starts in the bottom left corner and the resource budget was set to 50. The pink line indicates the path the agent took and the green dots represent successful drill sites. The two plots on the right show the trace of the Gaussian process covariance matrix and the RMSE of the beliefs with the true map. Note that for even comparison we evaluated $\text{Tr}(\Sigma)$ and the RMSE all using the same Gaussian process setup, even though MCTS-DPW was the only method to use Gaussian process beliefs online. The plots show the average and standard deviation from 50 different simulation runs with the resource budget set to 100.} \label{fig:rover_results}
 \vspace{-2mm}
\end{figure*}

\begin{table*}[t]
\resizebox{\textwidth}{!}{
\normalsize 
\begin{tabular}{ lcccccccccccccc }
\toprule
      &\multicolumn{4}{c}{$\sigma_{s}=0.1$} & &\multicolumn{4}{c}{$\sigma_{s}=0.5$} & & \multicolumn{4}{c}{$\sigma_{s}=1.0$} \\
    \cline{2-5}
    \cline{7-10} 
    \cline{12-15}\\
    Budget & POMCP & POMCP-GCB & MCTS-DPW & Raster & & POMCP & POMCP-GCB & MCTS-DPW & Raster & & POMCP & POMCP-GCB & MCTS-DPW & Raster \\
    \hline
    $30$  &  $0.80$  & $0.96$ & $\boldsymbol{3.42}$ & --   & & $0.62$ & $0.42$ & $\boldsymbol{1.88}$ & --  & & $0.64$ & $0.70$ & $\boldsymbol{1.54}$ & -- \\
    $60$  & $-0.04$ & $1.36$ & $\boldsymbol{4.84}$ & --   & & $0.22$ & $0.06$ & $\boldsymbol{2.56}$ & --  & & $0.18$ & $0.48$ & $\boldsymbol{1.78}$ & -- \\
    $100$ & $-0.50$  & $1.12$ & $\boldsymbol{5.60}$ & $2.24$ & & $0.32$ & $-1.50$ & $\boldsymbol{2.80}$ & $2.24$ & & $-0.68$ & $-0.06$ & $1.90$ & $\boldsymbol{2.24}$  \\
\bottomrule
\end{tabular}}
\caption{Comparison of the four methods as the resource budget and spectrometer measurement noise are varied. Each reported value is the average of 50 simulation runs.}
\label{table:rover_results}
\vspace{-7mm}
\end{table*}

\subsection{Rover Exploration}
We now introduce a new AIPPMS benchmark problem known as the Rover Exploration problem, which is inspired by multiple planetary rover exploration missions \cite{francis2017aegis, mangold2021perseverance, heldmann2016site}. The rover begins at a specified starting location and has a fixed amount of energy available to explore the environment and reach the goal location. The rover is equipped with a spectrometer and a drill. Drilling reveals the true state of the environment at the location the drill sample was taken and is a more costly action to take from a resource budget perspective. Conversely, the spectrometer provides a noisy observation of the environment and uses less of the resource budget. At each step, the rover can decide whether or not it wants to drill. The rover's goal is to collect as many unique samples as it can while respecting its energy constraints. The rover receives $+1$ reward for drilling a sample that it has not yet seen and $-1$ reward for drilling a sample that it has already previously collected. This AIPPMS problem highlights the importance of taking sensing actions to decrease the uncertainty in the belief-state before deciding to drill. 

For the Rover Exploration problem, we focus on the interplay between the resource budget allotted to the rover and the sensing quality of the spectrometer, where $\sigma_s$ denotes the standard deviation of a Gaussian sensor model. We also include a raster policy that attempts to fully sweep the environment in a raster pattern and evenly distribute its drilling actions along the way. However, the raster policy may not always be able to make it to the goal location in the specified budget in which case the rover receives $-\infty$ reward for a mission failure.

The environment is modeled as an $n \times n$ grid with $\beta$ unique measurement types in the environment with values between $0$ and $1$. To construct a spatially-correlated environment, we first start by sampling each grid cell value from an independent and identically distributed uniform distribution of the $\beta$ unique measurement types. Each cell is then averaged with the value of all of its neighboring cells. This process creates environments with spatial correlation as well as some random noise to simulate a realistic geospatial environment.

To implement the approach from Choudhury et al. on this benchmark problem, we maintain a discrete belief at each of the grid cells corresponding to the $\beta$ unique measurement types. When a measurement is received, the weights corresponding to the discrete values are updated based on a Gaussian sensor model. 

Fig. \ref{fig:rover_results} shows an example of a rover trajectory with the posterior mean and variance of the Gaussian process, as well as a comparison of $\text{Tr}(\Sigma)$ and the RMSE along the trajectory for the different methods. Table \ref{table:rover_results} summarizes the results as the resource budget and spectrometer sensing quality are varied. We highlight three key observations from the results in Fig. \ref{fig:rover_results} and Table \ref{table:rover_results}. 

First, as expected, MCTS-DPW is able to find more unique drill locations as the resource budget is increased. As the spectrometer noise is increased, less reward is received due to MCTS-DPW having less confidence in its belief about the environment. However, for POMCP and POMCP-GCB this is not the case. As the resource budget increases, we generally see that both POMCP and POMCP-GCB receive less reward since the agent has more opportunities to make mistakes by receiving a sample type it has already seen. MCTS-DPW is able to avoid this by effectively balancing exploration and exploitation by reasoning about its confidence in a particular region before drilling, leading it to make more risk-aware decisions. As the agent has drilled in more locations it will become more conservative in drilling in future locations unless it is very certain that the sample type is unique. Additionally, the Gaussian process belief enables reasoning about spatial correlation in the environment whereas the discrete belief supports reasoning only about the specific cell it has observed, but not about the relationships with other neighboring cells. This is one of the key reasons Gaussian processes are used for geospatial modeling.

Second, under significant amounts of noise ($\sigma_s = 1.0$), the deterministic raster policy is able to outperform MCTS-DPW. The raster policy simply tries to fully cover the space and evenly distribute its drill samples instead of attempting to reason online about the measurements it has received. It is important to note that the $\beta$ unique measurement types take on values between $0$ and $1$ so $\sigma_s = 1.0$ represents a very noisy signal. Therefore, in situations where the sensor is not reliable and the resource budget is sufficient, a deterministic raster policy may be the best option. However, often in real-world scenarios, the resource budget is not large enough to fully cover the area of interest and the onboard sensors have some degree of reliability.

Third, we see that MCTS-DPW is able to achieve the lowest $\text{Tr}(\Sigma)$ and RMSE along the trajectory, further emphasizing its ability to balance exploration and exploitation. MCTS-DPW consistently achieves the largest reward while simultaneously decreasing its uncertainty in the world belief and predicting the true world state the most accurately.


Finally, it is important to note that POMCP and POMCP-GCB are able to plan much faster than MCTS-DPW because they do not have to conduct inference over the Gaussian process belief online. There is a wide body of literature devoted to scaling Gaussian processes efficiently \cite{bruinsma2020scalable, damianou2013deep, snelson2007local} but in the scenarios where the agent is collecting $<1000$ samples the scalability of the Gaussian process proved not to be an issue in our experiments. In all our experiments, the average planning time for POMCP, POMCP-GCB, POMCPOW-GCB, and MCTS-DPW were on the order of 0.01, 0.02, 0.01, and 0.5 seconds, respectively.

\section{Conclusion}\label{sec:conclusion}
This work focused on the variant of the AIPP problem where the agent is equipped with multiple sensors. We formulated the AIPPMS problem as a belief Markov decision process with Gaussian process beliefs. Using an approximation of the mutual information along with the expected reward, we solved the belief MDP using MCTS-DPW. We showed that our approach was able to outperform those of previous AIPPMS approaches as well as the POMCPOW-GCB algorithm by balancing information-gathering with exploitative reward-seeking actions. Future work includes integrating more efficient Gaussian process belief representations, such as adaptive query resolutions and expanding the belief representation to reason over Gaussian process parameters. These areas of future research would allow a similar formulation to be extended to a variety of real-world problems.

\vspace{-2mm}
\section*{Acknowledgments}
We would like to thank Kyle Wray, Shushman Choudhury, Jordan Ott, and Lauren Ho-Tseung for their insightful discussions throughout the process of this work.




\printbibliography

\end{document}